\pdfoutput=1

\documentclass[11pt]{article}

\usepackage{acl}

\usepackage{times}
\usepackage{latexsym}

\usepackage{mathtools}
\usepackage{amsfonts}

\usepackage[T1]{fontenc}

\usepackage[utf8]{inputenc}

\usepackage{microtype}

\title{Finetuning a Kalaallisut-English machine translation system using web-crawled data}

\author{Alex Jones \\
  Dartmouth College \\
  \texttt{alexander.g.jones.23@dartmouth.edu}}

\begin{document}
\maketitle
\begin{abstract}

West Greenlandic, known by native speakers as Kalaallisut, is an extremely low-resource polysynthetic language spoken by around 56,000 people in Greenland. Here, we attempt to finetune a pretrained Kalaallisut-to-English neural machine translation (NMT) system using web-crawled ``pseudoparallel" sentences from around 30 multilingual websites. We compile a corpus of over 93,000 Kalaallisut sentences and over 140,000 Danish sentences, then use cross-lingual sentence embeddings and approximate nearest-neighbors search in an attempt to mine near-translations from these corpora. Finally, we translate the Danish sentences to English to obtain a synthetic Kalaallisut-English aligned corpus. Although the resulting dataset is too small and noisy to improve the pretrained MT model, we believe that with additional resources, we could construct a better pseudoparallel corpus and achieve more promising results on MT. We also note other possible uses of the monolingual Kalaallisut data and discuss directions for future work. We make the code and data for our experiments publicly available. \footnote{\url{https://github.com/AlexJonesNLP/KALComp}}

\end{abstract}

\section{Introduction}

Low-resource machine translation—translation involving languages with very few linguistic resources, especially parallel data—is a massive challenge in NLP. The lack of data makes it incredibly difficult to train and evaluate NMT systems and, to a lesser extent, rule-based and statistical MT systems. Several works have summarized the manifold challenges that come with attempting low-resource MT \citep{haddow-2021-lowres, ranathunga-2021-lowres}.

In this paper, we tackle MT involving Kalaallisut, a polysynthetic Eskimo-Aleut language spoken by around 56,000 people in Greenland. Only a couple works have attempted Kalaallisut-English MT specifically \citep{de_mol_2020, kelly_2020}, both of which see poor performance due to noisy and insufficient data. For this reason, we attempt different methods than them for aligning crawled data.

A small number of works have looked at MT and morphological segmentation for other polysynthetic Eskimo-Aleut languages, including Inuktitut and Yupik \citep{roest-etal-2020-machine, joanis-etal-2020-nunavut, ngoc-le-sadat-2020-revitalization, micher_2018, le_sadat_2020, micher-2018-using}. Among these efforts was the creation of an Inuktitut-English parallel corpus, the Nunavut Hansard corpus.

In this paper, we first web-crawl monlingual texts in Kalaallisut and Danish from multilingual Greenlandic websites. We then attempt to find ``pseudoparallel" sentence pairs, or sentences that are near-translations of each other, between these two sets of monolingual corpora. To this end, we deploy cross-lingual sentence embeddings and a highly optimized library for approximate nearest-neighbors search. We then try to finetune pretrained open-source MT model from HuggingFace \citep{wolf-etal-2020-transformers}, namely the Helsinki-NLP/OPUS-MT model\footnote{\url{https://huggingface.co/Helsinki-NLP/opus-mt-kl-en}} \citep{tiedemann-thottingal-2020-opus} trained using the MarianMT toolkit \citep{junczys-dowmunt-etal-2018-marian}.

\section{Methodology}

\subsection{Web crawling}
The first step in our process was to gather monolingual sentences from data sources containing ``similar" texts. In our case, we chose to crawl texts from multilingual websites that contain the same content in multiple languages. Originally, we planned to crawl texts in Kalaallisut, Danish, and English. However, due to the scarcity of English text we were able to obtain, we ended up not using the English sentences. The resulting set of Kalaallisut and Danish sentences is a ``comparable corpus": a collection of similar texts (i.e. about the same topics or events) in two or more languages. Intuitively, one might assume that such corpora sometimes contain sentences that are (near-)translations of each other, which is indeed the case \citep{schwenk-etal-2021-wikimatrix, schwenk-etal-2021-ccmatrix, bucc-2021-building}.

We compile a list of 29 Greenlandic websites with content in Kalaallisut and Danish. This included a diverse mix of news, government, corporate, scientific, and travel/hospitality websites. Next, we used the Python \texttt{requests}\footnote{\url{https://docs.python-requests.org/en/latest/}} library in tandem with \texttt{BeautifulSoup}\footnote{\url{https://www.crummy.com/software/BeautifulSoup/bs4/doc/}} to obtain and parse HTML from these sites. In order to maximize the amount of text we crawl, we scrape text from webpages recursively. That is, we scrape text from a webpage, and then we retrieve links on that page that direct to other pages on the website. This allows us to scrape large portions of certain websites. After scraping, we clean the text using the \texttt{cleantext}\footnote{\url{https://pypi.org/project/clean-text/}} library. We end up with 93,093 Kalaallisut sentences and 140,802 Danish sentences, after removing duplicates. 


\subsection{Bitext mining}

The bitext mining task consists of trying to find sentence pairs that are translations or near-translations of each other between two sets of monolingual sentences in different languages.

A large number of approaches have been attempted for this task over the past two decades \citep{zhao-vogel-adaptive, resnik-smith-2003-web, munteanu-etal-2004-improved, fung-cheung-2004-multi, munteanu-marcu-2006-extracting, azpeitia-etal-2017-weighted, azpeitia-stacc, BOUAMOR18.8, Hangya2018Unsupervised, schwenk-2018-filtering, ramesh-sankaranarayanan-2018-neural, artetxe-schwenk-2019-margin, artetxe-schwenk-2019-laser, hangya-fraser-2019-unsupervised, wikimatrix, ccmatrix, wu-etal-2019-machine, keung-etal-2020-unsupervised, tran-criss-2020, kvapilikova-etal-2020-unsupervised}. However, current state-of-the-art techniques generally involve embedding sentences in both languages using a single cross-lingual encoder and then performing approximate nearest-neighbors search with some similarity metric. We hew closely to the method introduced in \citet{artetxe-schwenk-2019-margin}.

\subsubsection{Cross-lingual sentence embeddings}

In order to create vector representations of sentences in different languages that are useful for the bitext mining task, we employ the cross-lingual sentence encoder LaBSE (\textsc{L}anguage-\textsc{a}gnostic \textsc{B}ERT \textsc{S}entence \textsc{E}mbedding) \citep{feng2020labse}. Cross-lingual encoders aim to create language-agnostic representations of sentences. In theory, two semantically equivalent sentences in different languages (i.e. translations) should map to the exact same vector. This makes them the ideal tool for this task.

LaBSE combines pretraining on the masked language modeling (MLM) and translation language modeling (TLM) \citep{conneau2019cross} tasks with the dual encoder translation ranking and additive margin softmax objectives. The model is trained on data in 109 languages, all of which have at least some parallel data aligned with English. LaBSE achieves state-of-the-art results when used for the bitext mining task \citep{feng2020labse, reimers-gurevych-2020-making}. We use the implementation of LaBSE available on the \texttt{Sentence Transformers} library.\footnote{\url{https://www.sbert.net}} \citep{reimers-gurevych-2019-sentence}. The embeddings are 768-dimensional, and the embedding is done with a Quadro RTX 6000 GPU.

\subsubsection{Dictionary translation}

The issue with using LaBSE for embedding Kalaallisut sentences is that it was not trained on any Kalaallisut data. Furthermore, LaBSE doesn't even have training data for languages in the Eskimo-Aleut family that Kalaallisut is a part of, meaning it cannot leverage related-language data for zero-shot transfer as it could in less resource-scarce scenarios. Because of this, it is desirable to translate all or part of the Kalaallisut text to a higher-resource language before attempting similarity search with LaBSE.


We end up using the rather rudimentary method of translating parts of the Kalaallisut text using Kalaallisut-English and Kalaallisut-Danish dictionaries\footnote{\url{https://www.mobileread.com/forums/showthread.php?t=20480&page=11}}, resulting in a type of ``code-switched" corpus. However, these dictionaries were highly incomplete, so we were only able to translate $\approx 16\%$ of the Kalaallisut words to English or Danish.

\subsubsection{BPE}

Kalaallisut is a highly polysynthetic language, meaning it has many morphemes per word on average and a huge (effectively limitless) vocabulary size. This poses a significant challenge for a host of NLP problems, including cross-lingual similarity search and MT. 

\citet{de_mol_2020} explored various methods for morphological segmentation of Kalaallisut, as well as their downstream consequences on MT. Although Conditional Random Fields performed best on a gold-standard morphological segmentation evaluation in their papere, we elect to use the simpler byte-pair encoding (BPE) algorithm \citep{sennrich-etal-2016-neural}, a language-agnostic and non-linguistic tokenization scheme that produces subword units based on statistical analysis of a training corpus. BPE has proven to be useful in MT training and has become a staple tokenization method in NLP. Segmentation has also been shown to help bitext mining performance between polysynthetic and analytic languages, such as Inuktitut and English \citep{jones-etal-2021-massively}.

We use the BPE implementation from the Python package \texttt{youtokentome}\footnote{\url{https://pypi.org/project/youtokentome/}}, with a vocabulary size of 10,000 and otherwise default settings. We train the BPE model on all the monolingual Kalaallisut data.

\subsubsection{Approximate KNN search}
\paragraph{Similarity search}

We use the \texttt{Faiss} (\textsc{F}acebook \textsc{A}I \textsc{S}imilarity \textsc{S}earch) library \citep{faiss} to perform similarity search between LaBSE embeddings of Danish and code-switched Kalaallisut sentences. \texttt{Faiss} is an aggressively optimized package that allows for extremely fast $k$-nearest neighbors search on GPUs. We use a Python implementation\footnote{\url{https://pypi.org/project/faiss-gpu/}} of \texttt{Faiss}, although the original package\footnote{\url{https://github.com/facebookresearch/faiss}} is written in C++. We use $k=4$ neighbors and a batch size of $100$ for our hyperparameter settings, and run the search on a Quadro RTX 6000 GPU. We also run both forward and backward searches, i.e. $\forall x \in \mathcal{X}$ we find the most similar sentence $y \in \mathcal{Y}$ and vice-versa. We then take the union of both searches to give us a larger pool of candidate sentence pairs to choose from (see \citet{artetxe-schwenk-2019-margin} for more details).

\paragraph{Margin criterion}

Many works use the cosine similarity metric to quantify similarity between vector representations, which is simply the dot product of two normalized vectors. However, we use the slightly more sophisticated margin criterion from \citet{artetxe-schwenk-2019-margin} instead, which has been shown to give better performance on the bitext mining task compared to simple cosine similarity. The margin score between two sentence embeddings $x \in \mathcal{X}$ and $y \in \mathcal{Y}$ is given by:
\begin{equation}\label{eq:margin_score}
\begin{split}
&\mathrm{score_{margin}}(x,y) = \\
&\frac{2k\cos{(x,y)}}{\sum_{z\in NN_{k}(x)}{\cos{(x,z)}}+\sum_{z\in NN_{k}(y)}{\cos{(y,z)}}}\nonumber
\end{split}
\end{equation}
where $NN_{k}{(v)}$ denotes the set of $k$-nearest neighbors of vector $x \in \mathbb{R}^n$ and $\cos(x,y)$ is the cosine similarity between $x$ and $y$, i.e. $\frac{x \cdot y}{\|x\|\|y\|}$. The margin score measures the ratio between the cosine similarity of $x$ and $y$ and the average cosine similarity of $x$ and its $k$-nearest neighbors and $y$ and its $k$-nearest neighbors. The vector pair $(x, y)$ that maximizes the margin criterion is the one that ``stands out" the most in terms of similarity among its neighbors.

We use a margin score threshold of $1.04$ to extract our final set of pseudoparallel sentence pairs, identical to \citet{schwenk-etal-2021-wikimatrix}. There is obviously a precision-recall tradeoff in selecting a margin threshold, and slightly higher thresholds such as $1.06$ have also shown success \citep{schwenk-etal-2021-ccmatrix}. Since our monolingual datasets are so small already, we opt for a threshold that will give us somewhat more sentence pairs at the expense of quality. We end up with 6,393 sentence pairs.

\subsubsection{Post-translation}

Recall that we performed similarity search between embeddings of Danish sentences and code-switched Kalaallisut sentences. After mining and filtering with a margin threshold, we simply map the code-switched Kalaallisut sentences back to the ``pure" Kalaallisut sentences they came from. This resulted in a pseudoparallel Kalaallisut-Danish corpus. However, the model we wished to finetune is a Kalaallisut-English MT system. So we translate the Danish sentences to English using the Helsinki-NLP/OPUS-MT Da-En model\footnote{\url{https://huggingface.co/Helsinki-NLP/opus-mt-da-en}}. We run this model on a NVIDIA TITAN V GPU, translating in batches of $32$ sentences. The resulting training set is a synthetic Kalaallisut-English pseudoparallel corpus.

\subsection{NMT finetuning}

The next step was to finetune the pretrained Kl-En MT model on the pseudoparallel data we had gathered. We finetune the model for only one epoch with a batch size of $8$, for a total of $720$ optimization steps. We use a learning rate of 2e-5 and a weight decay parameter of $0.01$. Unfortunately, due to technical difficulties we were forced to run our code in a Google Colab instance, which meant using a Tesla K80 GPU instead of a more powerful GPU. Finetuning took around $23$ minutes.

\section{Evaluation}

We evaluate using BLEU score \citep{papineni-etal-2002-bleu}, the most popular metric in the MT literature. The test set we use is the same set\footnote{\url{https://object.pouta.csc.fi/OPUS-MT-models/kl-en/opus-2020-01-09.test.txt}} the pretrained model was evaluated on, which consists of texts from the JW300 parallel corpus \citep{agic-vulic-2019-jw300}. Disappointingly, the finetuned model achieves a BLEU score of only $14.67$, compared to a score of $27.75$ for the pretrained model. There are multiple factors that may explain this. For one, the pseudoparallel data we extracted is incredibly noisy, despite attempts to bootstrap similarity search using dictionary translation. A glance at the dataset reveals that many, if not most, of the sentence pairs are not translations or even near-translations. Second, the pretrained model is almost certainly overfit to the JW300 data, since this corpus is also the source of the training data. Earlier in the experiment, when we tried to translate the crawled Kalaallisut to English, the pretrained model generated gibberish sentences that seemed to be replicating what it saw in the training corpus, with apparently no connection to the input. So although the data we mined is undoubtedly noisy, we do not put much stock in this evaluation due to the problem of overfitting to the test set's underlying data distribution. Future efforts should seek to use an unbiased test set, which we were unfortunately unable to find.

\section{Discussion}

It is unfortunate that we were unable to mine a sufficient number of high-quality sentence pairs to improve a pretrained MT system. However, we would like to take stock of the contributions of this project, as well as discuss avenues for potential improvement. 

Although our efforts did not result in a high-quality aligned corpus, we still managed to construct a Kalaallisut-Danish \textit{comparable} corpus purely from web-crawled data. We plan to release this data publicly for future research endeavors. Furthermore, the nearly 100K Kalaallisut sentences we extracted could be used for unsupervised tasks such as language modeling.

We believe we could improve the bitext mining process for Kalaallisut using larger dictionaries with better coverage, and/or a decent pretrained MT system. A larger Kalaallisut-English dictionary appears to be in the works\footnote{\url{https://scholarspace.manoa.hawaii.edu/handle/10125/26190}}, but we couldn't locate any text files associated with it. There is a rule-based Kalaallisut-Danish MT system\footnote{\url{https://nutserut.gl/en}} hosted by Oqaasileriffik, the language secretariat of Greenland. Cursory experiments with this system yielded promising results, but we were unable to co-opt the system for batch translation (e.g. in a Python environment) due to the fact that this tool is currently just an on-demand web API and UI. In the future, we can contact the makers of this tool to see if we can use it for large-scale translation.

Our crawling pipeline was also suboptimal due to its uniformity (and other reasons, such as time constraints and Python recursion depth limits), and we could likely mine considerably more monolingual sentences using custom parsers. This should lead to improved downstream performance as well.

\section{Conclusions}

In this paper, we attempt to finetune a pretrained Kalaallisut-English MT system using web-crawled data. We scrape monolingual text from roughly $30$ multilingual websites in Kalaallisut and Danish and then try to mine pseudoparallel sentences using LaBSE and \texttt{Faiss}, after which we translate the Danish to English to create a synthetic, pseudoparallel Kalaallisut-English corpus. Although finetuning results are disappointing, we offer the Kalaallisut-Danish comparable corpus as a novel and valuable resource, and suggest multiple directions for possible improvement.

\bibliography{anthology,custom}

\end{document}